\documentclass[journal]{IEEEtran}

\usepackage{url}
\usepackage{color}
\usepackage{amsmath}
\usepackage{array,graphicx,subfigure}
\usepackage{multirow}

\usepackage{algorithm}
\usepackage{algpseudocode}

\newcommand{\tabincell}[2]{\begin{tabular}{@{}#1@{}}#2\end{tabular}}

\newcommand{\doublecheck}[1]{\textcolor{blue}{#1}}
\newcommand{\keypoint}[1]{\vspace{0.0cm}\noindent\textbf{#1}\quad}
\newcommand{\cut}[1]{}

\newcommand{\sbvr}{SBVR}
\newcommand{\fgsbvr}{FG-SBVR}
\newcommand{\moduleName}{relation module}
\newcommand{\ModuleName}{Relation Module}

\newcommand{\todo}{\textcolor{red}{TODO}}

\newcommand{\ie}{\textit{i}.\textit{e}.}
\newcommand{\eg}{\textit{e}.\textit{g}.}
\newcommand{\etc}{\textit{etc}}

\ifCLASSINFOpdf
  
\else
 
\fi

\begin{document}

\title{Fine-Grained Instance-Level Sketch-Based \\ Video Retrieval}

\author{Peng~Xu,~Kun~Liu,~Tao~Xiang,~Timothy~M.~Hospedales,~Zhanyu~Ma,~Jun~Guo,~and~Yi-Zhe~Song
\thanks{Peng Xu is with School of Computer Science and Engineering, Nanyang Technological University, Singapore. E-mail: peng.xu@ntu.edu.sg~Homepage:~http://www.pengxu.net/   GitHub:~https://github.com/PengBoXiangShang}
\thanks{Kun Liu, Zhanyu Ma, and Jun Guo are with Beijing University of Posts and Telecommunications, China. E-mail: \{liu\_kun, mazhanyu, guojun\}@bupt.edu.cn }
\thanks{Tao Xiang, Yi-Zhe Song are with Centre for Vision, Speech and Signal
Processing (CVSSP), University of Surrey, United Kingdom. E-mail: \{t.xiang, y.song\}@surrey.ac.uk }
\thanks{Timothy~M.~Hospedales is with University of Edinburgh, United Kingdom. E-mail: t.hospedales@ed.ac.uk }
}


\maketitle

\begin{abstract}
Existing sketch-analysis work studies sketches depicting static objects or scenes. In this work, we propose a novel cross-modal retrieval problem of fine-grained instance-level sketch-based video retrieval (FG-SBVR), where a sketch sequence is used as a query to retrieve a specific target video instance. Compared with sketch-based still image retrieval, and coarse-grained category-level video retrieval, this is more challenging as both visual appearance and motion need to be simultaneously matched at a fine-grained level.  We contribute the first FG-SBVR dataset with rich annotations. We then introduce a novel multi-stream multi-modality deep network to perform FG-SBVR under both strong and weakly supervised settings. The key component of the network is  a \moduleName{}, designed to prevent model overfitting given scarce training data.  We show that this model significantly outperforms a number of existing state-of-the-art models designed for video analysis.
\end{abstract}

\begin{IEEEkeywords}
fine-grained video retrieval, sketch-based video retrieval, sketch dataset, cross-modal matching, triplet ranking, meta-learning inspired techniques.
\end{IEEEkeywords}

\IEEEpeerreviewmaketitle

\section{Introduction}

\IEEEPARstart{I}{t} 
is said that one sketch speaks for a hundred words. Sketch provides a convenient abstraction to bridge  concepts and pixels, via capturing both salient detail and topology.  Sketch is now convenient and widely captured given the modern prevalence of touch-screen devices. This has led to a flourishing~\cite{xu2020deep} of sketch-related research in recent years including sketch-based image retrieval~\cite{yu2016sketch,xu2018cross,collomosse2019livesketch}, sketch-generation~\cite{ha2017sketchrnn},  segmentation \cite{schneider2016sketchSegmentation},   \cut{sketch-related zero-shot learning \cite{hu2018classifier},}  hashing~\cite{xu2018sketchmate},  abstraction~\cite{muhammad2018learning}, scene understanding~\cite{ye2016human,xie2019deep}, and self-supervised representation learning~\cite{xu2020deepself}. However, all of these studies only work with sketches depicting static objects or scenes, and analysis of sketching of motion is much under-studied. 

\begin{figure}[t]
	\centering
		\includegraphics[width=\columnwidth]{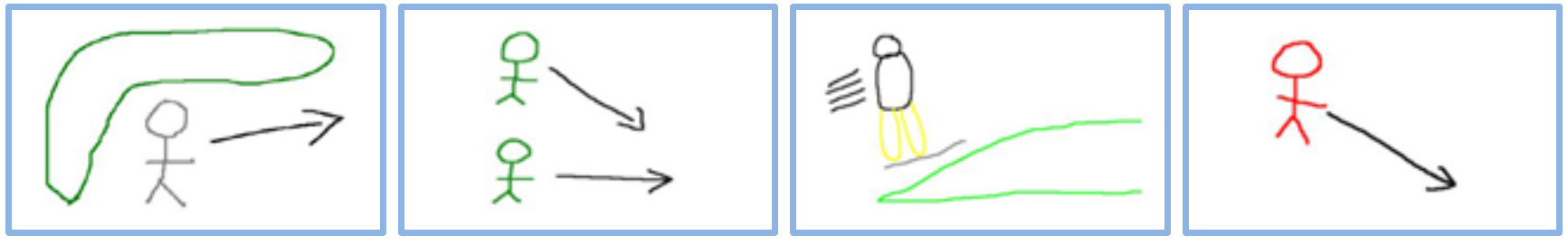}

	\caption{Sketch examples of the existing SBVR dataset~\cite{collomosse2009storyboard}.}
	\label{fig:john_dataset}
\end{figure}

Humans recall and describe events from episodic memory~\cite{tulving1985elements} with selective effects~\cite{madore2018selective}. Visual recollections mainly contains the appearance and actions of key objects (\eg, movements, spinning, rising). Combined with free-hand drawing of arrows or lines, sketch can simultaneously describe the appearance and motion of objects corresponding to such typical human recollections. Motivated by this potential, sketch-based video retrieval (\sbvr) was first proposed in \cite{collomosse2009storyboard}, and followed up in several subsequent studies \cite{hu2010motion,hu2012annotated,hu2013markov,james2014interactive}. However, these early methods are relatively \textit{coarse-grained}, with sketched objects providing almost symbolic category indicators, rather than fine-grained details where sketch really shines as an alternative to conventional tagging approaches \cite{sangkloy2016sketchy,yu2016sketch}. Moreover the associated datasets are not large enough to train contemporary deep methods, and are not instance-level, in the sense of there being a set or videos rather than a single target video for each query sketch.
Some examples of the TSF dataset introduced in \cite{collomosse2009storyboard} are shown in Figure~\ref{fig:john_dataset}, in which objects are \emph{iconic}, without any fine-grained appearance information. This fails to exploit the full expressiveness of sketch and undermines the practical motivation for SBVR, since conventional symbolic tags (`person') could be a more convenient query modality there.


\begin{figure}[!t]
\begin{center}
\includegraphics[width=\columnwidth]{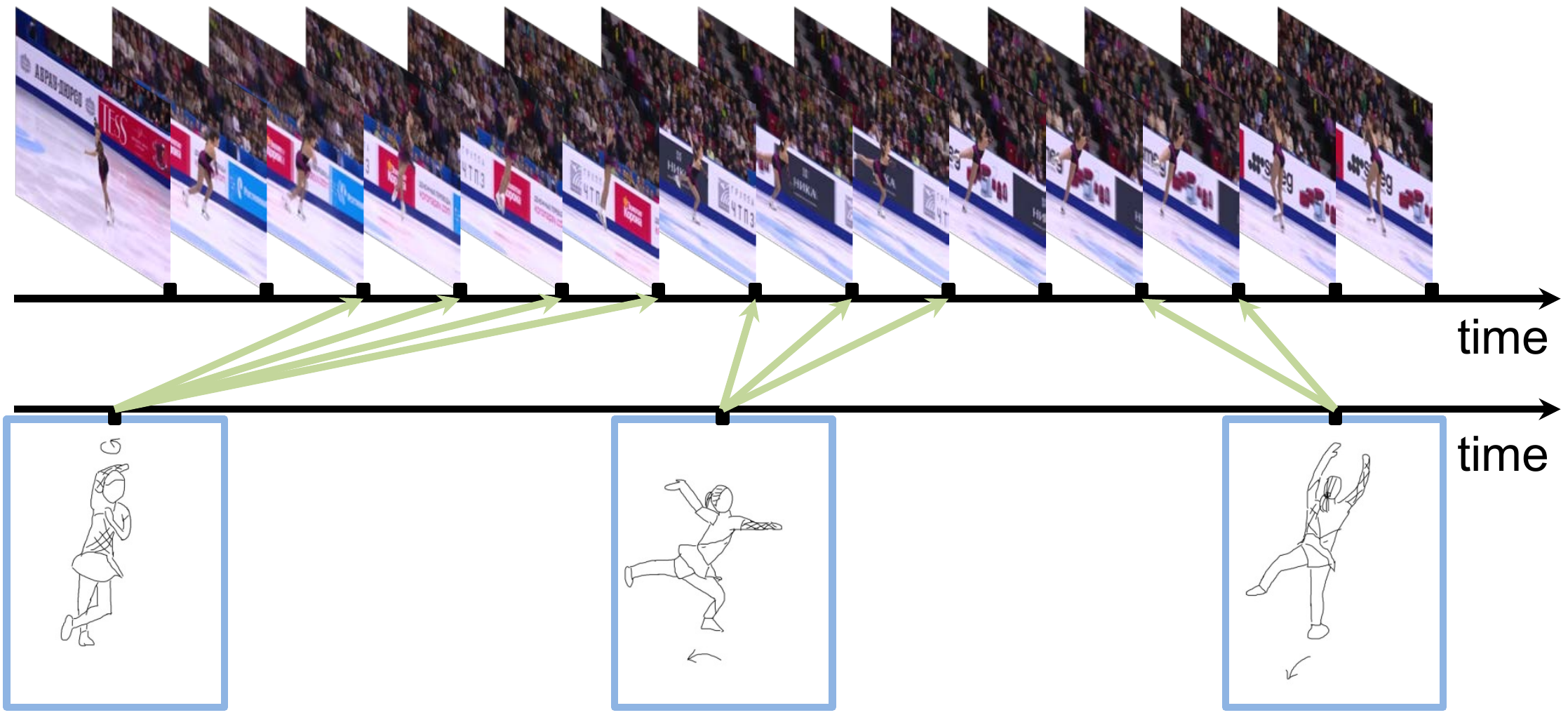}
\end{center}
   \caption{Illustration of fine-grained instance-level sketch-based video retrieval. A sketch-sequence (bottom) is connected to the video frames they summarise (top). Best viewed in color.}
\label{fig:film}
\end{figure}

In this paper we provide the first study of genuinely fine-grained instance-level sketch-based video retrieval~(\fgsbvr). This task is extremely challenging since it not only needs to solve all the difficulties common to static-sketch cross-modal retrieval (\ie, matching abstract and sparse line-drawings to dense pixel renderings of perspective projections), but also requires understanding of motion depiction in sketch, and registering sketches to specific time windows within a temporally extended video. To support research in this area, we introduce the first {\fgsbvr} dataset containing $1,448$ sketches corresponding to $528$ figure skating video clips. Going beyond previous studies: (i) All sketches depict subtle  appearance and pose details of skaters including body posture, hand gesture, clothing, and hair style. (ii) Skater movements are summarised by fine-grained motion vectors indicating skater glissades, spins and jumping. (iii) To represent temporally extended video rather than instantaneous actions, video clips are described by multi-page sketches. One example is illustrated in Figure~\ref{fig:film}. This multi-page ``skater + motion vector'' sketch format means that a successful  {\fgsbvr} model must solve  three challenging matching/alignment issues: (i) fine-grained visual appearance matching between drawn skaters and video image frames, (ii) fine-grained motion matching between sketched motion indicators and the video motion, (iii) alignment of pages within the sketch query to sub-sequences within the video. 

In order to solve this cross-modal matching problem, we introduce the first deep model for {\fgsbvr}. It is a multi-stream multi-modality deep neural network. Specifically, the network is designed to align the video and sketch-sequence modalities in a joint embedding space so that they can be compared by a specific distance. Taking into consideration the unique ``skater + motion vector'' sketch format, each modality is modelled by a sub-network composed of a ``appearance'' stream and  a ``motion'' stream. Within each stream, cross-modal matching is modelled by triplet ranking.  
\textcolor{black}{This network design is against the recent trend in video analysis dominated by 3D convolutional networks \cite{wang2018non,hara2018can,liu2018t}, 
which do not explicitly separate  dynamic and static video content.} We found that with the large modality gap between video and sketch and the scarce training data, decomposing the dynamic and static aspects of both modalities explicitly becomes crucial. 

To further address the training data scarcity problem,  inspired by existing meta learning based few-shot learning work~\cite{sung2018learning,xie2018comparator}, we introduce a \moduleName{} into our {\fgsbvr}. One of the most effective ways of improving generalization by meta learning is training a non-linear comparison -- or relation -- module. 
The \moduleName{} improves the learned representation by modelling the non-linear relationship between sketch-clip pairs, and benefits from learning from more negative pairs compared to triplet loss alone. Our {\fgsbvr} can be thus trained effectively
even with  sparse data. Furthermore, we explore both the strongly supervised setting (using ground-truth sketch page-frame alignment annotation during training), and the weakly-supervised learning (no within-video sketch-frame correspondence) setting based on multi-instance learning (MIL). 

Our main contributions can be summarised as: (i) We propose the novel and challenging problem of fine-grained  sketch-based video retrieval. (ii) We contribute the first {\fgsbvr} dataset with extensive ground truth annotation\footnote{Our dataset and code will be made public.}. (iii) We develop a novel multi-stream multi-modality deep network to solve  {\fgsbvr} by explicitly decomposing appearance and motion. A \moduleName{} is also introduced in the network to prevent overfitting. (iv) We explore learning this framework with both strong- and weak supervision using multi-instance learning. 
Extensive experiments are conducted to show that the proposed model outperforms a number of state-of-the-art video analysis baselines.  

\section{Related Work}

\noindent\textbf{Video Retrieval}\quad
Many video retrieval techniques are query-by-example (QBE) \cite{tan1999framework}, in which users provide
(visual, textual, audio, \etc) examples of the content they seek. According to query modality, video retrieval spans \cite{li2015face,araujo2018large}:
(i) image-to-video (I2V) retrieval \cite{araujo2018large}, (ii) text-to-video (T2V) retrieval \cite{xu2015jointly, mithun2018learning}, and
(iii) video-to-video (V2V) retrieval, \eg, via hashing \cite{coskun2006spatio,li2012robust,ye2013large,chen2018nonlinear,song2018self}. V2V is a unimodal task,  T2V is a cross-modal task, and I2V is somewhere between.
However, all these query modalities have some drawbacks: (i) Image query only provides  appearance for one moment without  dynamic clues. 
(ii) Text generally needs a lot of words or sentences (for example, the appearance, movement and relative position of objects) if the video is to be described in detail, rather than just used as a list of keyword tags. (iii) Video query matches the modality to the data to retrieve, but it may be difficult to obtain a representative video that matches the desired video to be retrieved. 

\noindent\textbf{Sketch-Based Video Retrieval}\quad The pioneering work \cite{collomosse2009storyboard}  on {\sbvr} proposed a probabilistic model for content-based video retrieval driven by free-hand sketch queries depicting both objects and their movement (via dynamic cues, \ie, lines and arrows). This work led to a series of subsequent research including tracking visual key-points in videos to form short trajectories which can be clustered to form tokens summarising video content. These can then be matched to a color and motion description of a query sketch by a Viterbi-like process \cite{hu2010motion}. This method was  improved in~\cite{hu2012annotated} as a hybrid ``semantic sketch'' based video retrieval system by fusing the semantics of text with the expressiveness of sketch. Similarly, a Markov Random Field (MRF) optimisation based {\sbvr}  approach is proposed in~\cite{hu2013markov}, which combines shape, motion, colour and semantics within a single {\sbvr}  framework. Then, an index-based hybrid SBVR system is proposed in~\cite{james2014interactive}.
However,  the mentioned {\sbvr}   approaches have several drawbacks: (i) Their retrievals are relatively coarse-grained (see Figure~\ref{fig:john_dataset}). 
This undermines the unique practical advantage of sketch: to convey cues that are hard to describe with simple symbolic tags \cite{sangkloy2016sketchy,yu2016sketch}. Existing video tagging systems already allow text/tag-based video search for such coarse concepts.  (ii) Besides studying simple sketches, existing {\sbvr} methods address retrieving relatively instantaneous actions (see Figure~\ref{fig:john_dataset}). They do not address the challenge of retrieving actions with temporally extended structure.

\noindent\textbf{{\sbvr} Datasets}\quad There are only a few coarse-grained SBVR datasets~\cite{collomosse2009storyboard} to date. As mentioned earlier, in addition to the lack of \emph{instance-level} sketch-video  pairing, all sketches of these datasets are overly iconic object contours with motion lines and arrows. To facilitate {\fgsbvr} research, we therefore introduce the first fine-grained SBVR dataset. 
The unique features of our dataset are: (i) It contains fine-grained instance-level {\sbvr} data with one ground-truth video match to each sketch. (ii) It supports evaluation of more complex temporal logic in {\sbvr}, by allowing more than one page of action to describe a whole temporally extended video. (iii) It contains fine-grained visual detail enabling pose, clothing, and hairstyle to be used as matching cues. 

\noindent\textbf{Video Analysis Models}\quad Video analysis (\eg, action recognition) methods mainly include  RNN plus CNN \cite{donahue2015long,BiDiRNN2016}, two-stream networks \cite{simonyan2014two,wang2016temporal}, and
3D convolutional networks (\eg, C3D~\cite{tran2015learning}, P3D~\cite{qiu2017learning}, I3D~\cite{carreira2017quo}, T-C3D~\cite{liu2018t}, 3D ResNet18~\cite{hara2018can}, ARTNets~\cite{wang2018appearance}, Non-Local Neural Network~\cite{wang2018non}), STC-Net~\cite{diba2018spatio}, S3D~\cite{xie2018rethinking}, and MFNet~\cite{chen2018multi}). As mentioned earlier, we argue that the two-stream architecture is particularly suited for our multi-steam multi-modality alignment task because the dynamic and static streams of the video naturally correspond to the motion vector and static skater parts of sketches, respectively. This is validated in our experiments (see Section~\ref{sec:results}).  Our model is related to  fine-grained sketch-based image retrieval (FG-SBIR) models \cite{yu2016sketch,sangkloy2016sketchy}, but deals with a more challenging multi-stream matching problem. Importantly, none of the existing video analysis or FG-SBIR models exploit a \moduleName{} to address the overfitting problem due to data scarcity. 



\begin{figure*}[!t]
	\centering
		\includegraphics[width=\textwidth]{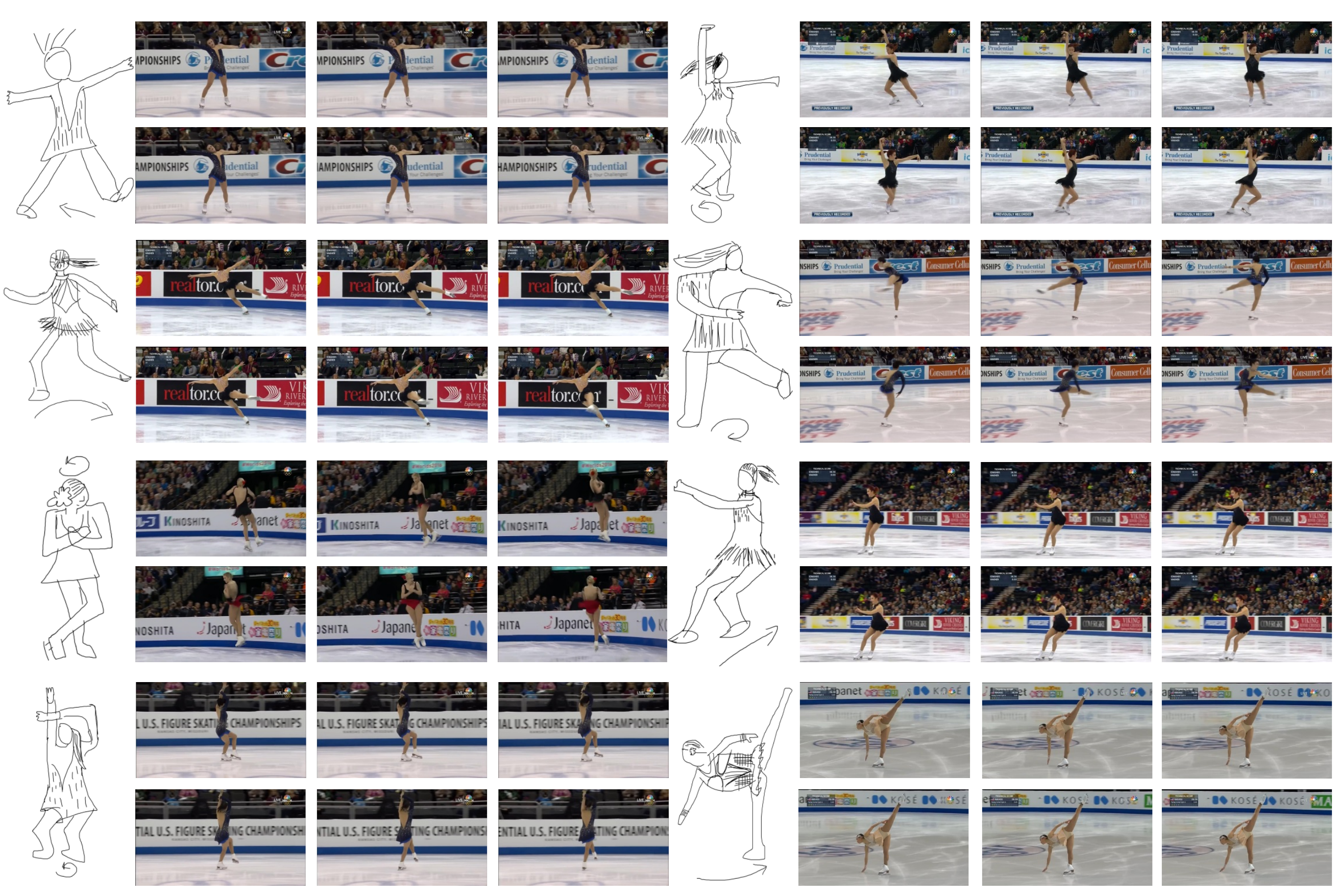}

	\caption{Examples of our figure skating FG-SBVR dataset. For each sketch page, its corresponding video frames (6 frames are selected) are shown. }
	\label{fig:our_samples}
\end{figure*}

%

\section{Fine-Grained Instance-Level SBVR Dataset}
\label{sec:dataset}

We contribute the first fine-grained instance-level sketch-based video retrieval dataset. It contains $528$ HD figure skating  clips and $1,448$ corresponding sketches. Our sketches contain both fine-grained appearance information, local dynamics, and longer-time frame dynamics via sequences of sketch `pages'. Some examples are shown in Figure~\ref{fig:our_samples}. Compared to sketches in prior datasets (Figure~\ref{fig:john_dataset}), ours have significantly more fine-grained details. We next describe the data collection and annotation process, and provide some quantitative comparisons with prior datasets.


\subsection{Data Collection}

\begin{figure*}[!ht]
	\centering
	\subfigure[]{
		\label{fig:video_length_hist}
      \includegraphics[width=0.22\textwidth]{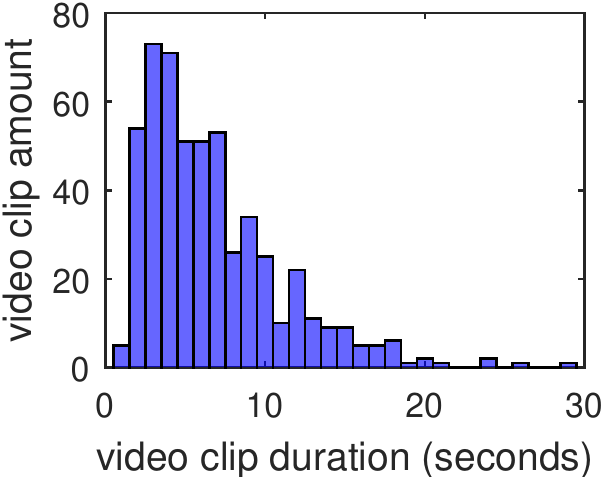}}
    \subfigure[]{
		\label{fig:how_many_svg_per_video_hist}
      \includegraphics[width=0.225\textwidth]{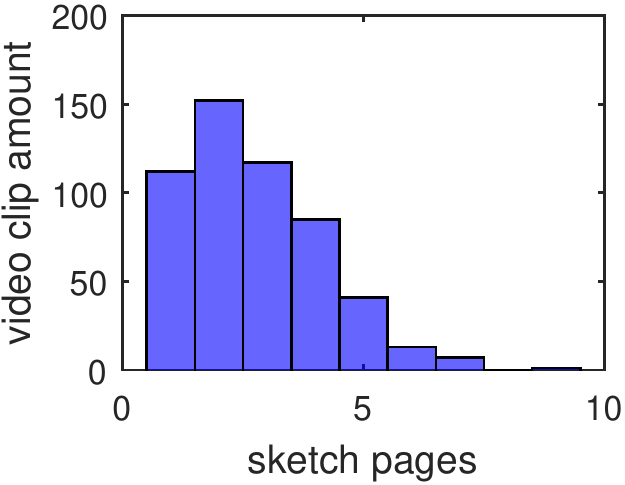}}
    \subfigure[]{
		\label{fig:how_many_stroke_hist}
      \includegraphics[width=0.23\textwidth]{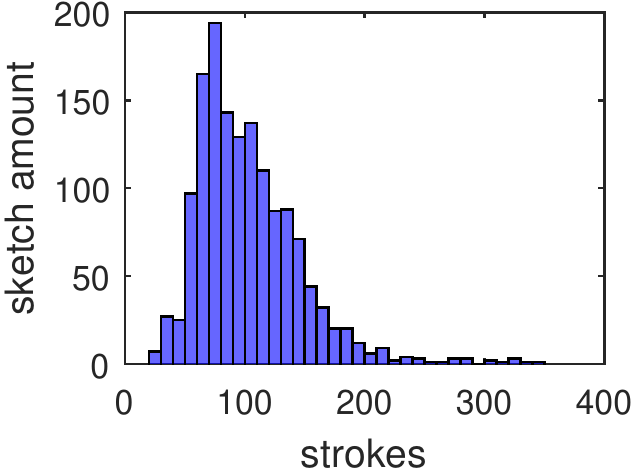}}
    \subfigure[]{
		\label{fig:scatter528}
      \includegraphics[width=0.21\textwidth]{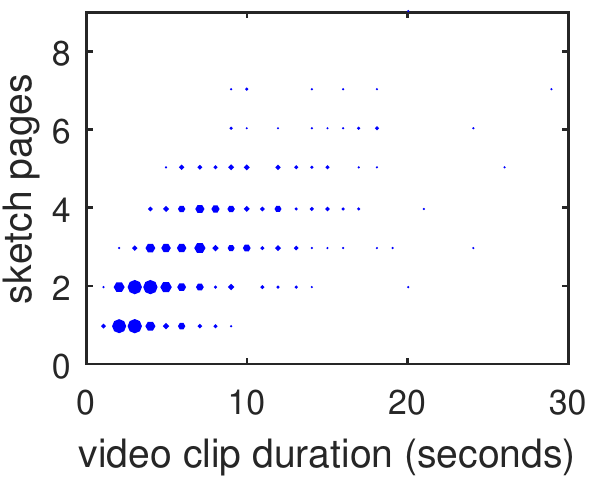}}
	\caption{Statistical analysis of our {\fgsbvr} dataset.}
	\label{fig:statistic_analysis_4_dataset}
\end{figure*}

\noindent\textbf{Videos}\quad We download diverse professional figure skating competition videos (\eg, US National, European and World Championships) from YouTube. From these, we selected $49$  female figure skating videos (duration: $6$ to $56$ minutes). For each video, both $720$P and $1080$P files are stored at $30$ FPS including audio channels with English narratives. The audio channel can support future research in speech or text modalities (\eg, extracting the keywords from narratives as `attribute vectors' describing the video). \textcolor{black}{We recruited $5$ skating fans to select representative clips from the original long videos.}
We cut out $528$ clips, with a total duration of $3,546$~seconds. The average length is $6.7$  seconds, with minimum $1$ and maximum $29$ seconds. Detailed duration statistics  are shown in  Figure~\ref{fig:video_length_hist}.

\noindent\textbf{Sketches}\quad The second step is to sketch the collected videos. We recruited $17$ skating fans who are amateur sketchers to sketch the clips. As can be seen in Figure~\ref{fig:our_samples}, due to lack of prior art training, these sketches are representative of the drawing abilities of the general population. 
The volunteers have a warm-up exercise, and then for each video clip, the volunteer can watch it several times and sketch what he/she has seen on a tablet, using their fingers or tablet stylus. Following recent practice \cite{eitz2012humans,sangkloy2016sketchy}, our sketches are saved in Scalable Vector Graphic (SVG) format that stores spatio-temporal stroke information. \cut{We also store one time vector for each sketch, in which the relative time sequence in milliseconds since the first point of sketch is recorded.}

Each sketch contains two parts: the skater depicted at certain posture representing a key moment of the routine, and a motion vector summarising the movements of the skater centred around that key moment. The motion vector is abstract and subjective, so some instructions are necessary to avoid some completely random interpretations. In particular, the volunteers were told that: (i) If a video clip just shows a static scene (\eg, skater keeps a static posture), draw the skater without any motion vectors. (ii) For jumping, the vector should be drawn above the skater's head. \textcolor{black}{For gliding and spinning, the vector should be drawn below the skater's feet.} (iii) Only one motion vector can be drawn within one sketch. (iv) One principle~\cite{miller1956magical} of information processing theory states that short-term memory is organised as chunks of meaningful units. \textcolor{black}{Thus,} the volunteers are free to decide how-many pages of sketch to use to summarise the video clip. 

\begin{table}[!tbp]
\caption{Comparison with previous sketch datasets. \textcolor{black}{``-'' means no stroke information.}}
\label{table:datasets_comparison}
\small

\begin{center}
\resizebox{\columnwidth}{!}{
\begin{tabular}{l | c | c | c | c | c | c}
\hline
\multirow{2}{*}{} & \multicolumn{4}{c|}{Strokes} & \multirow{2}{*}{\tabincell{c}{Resolution \\ ($W\times H$)}} & \tabincell{c}{Sketch \\ Amount} \\
\cline{2-5}
   & min & max & mean & std &  & \\
\hline
\hline
TU-Berlin~\cite{eitz2012humans} & 1 & 318 & 17.55 & 16.83 & $800 \times 800$    & 20k \\
Sketchy~\cite{sangkloy2016sketchy} & 1 & 434 & 17.91 & 16.06 & $640 \times 480$     & \textcolor{black}{75k}\\

QMUL Shoe & - & - & - & - & $256 \times 256$     & 419\\
QMUL chair & - & - & - & - & $256 \times 256$     & 297\\
QMUL handbag & - & - & - & - & $256 \times 256$     & 568\\
QMUL Chair-V2 & 1 & 138 & 12.79 & 9.84 & $800 \times 800$      &

 \textcolor{black}{1,275} \\
\hline
\hline
Ours & 26 & 345 & 102.40 & 43.47 & $768 \times 1024$ & 1,448\\
\hline
\end{tabular}
}
\end{center}

\end{table}

Among the $1,448$ sketches, there are $1,384$ motion sketches (with motion vectors, containing different kinds of spinning, jumping, \etc) and $64$ static sketches (without motion vectors). The sketch sequences range from one to nine pages. 
 Interestingly, $520$ ($\approx98\%$) of our video clips use less than seven sketches, which is consistent with human short term memory capacity being limited to seven chunks \cite{miller1956magical}. 
On average $2.7$ sketches are used to describe each video clip, and detailed statistics are shown in  Figure~\ref{fig:how_many_svg_per_video_hist}. 
In Figure~\ref{fig:scatter528}, we present a scatter plot to show the relation between video duration and corresponding sketch sequence, in which the radius of each point is proportional to the number of video clips. We can see that people  draw more sketches as video duration increases.

\noindent\textbf{Comparison with Other Datasets}\quad Compared with  existing sketch datasets, our dataset contains more details as reflected by the greater number of strokes~(see Figure~\ref{fig:how_many_stroke_hist}). This comparison is made quantitatively in Table~\ref{table:datasets_comparison}. It shows that our dataset contains similar number of sketches as previous single category sketch based cross-modal retrieval datasets (\ie, all except TU-Berlin~\cite{eitz2012humans} for sketch recognition and Sketchy~\cite{sangkloy2016sketchy} with 125 categories and 600 sketches per category). Drawing sketches with a reference image/video is very tedious making collecting large-scale datasets extremely difficult. Designing models that can be learned effectively with scarce data is thus a common challenge in sketch-based retrieval tasks.   



\subsection{Data Annotation}
\label{sec:data_annotation}

\begin{figure*}[!htp]
	\centering
		\includegraphics[width=\textwidth]{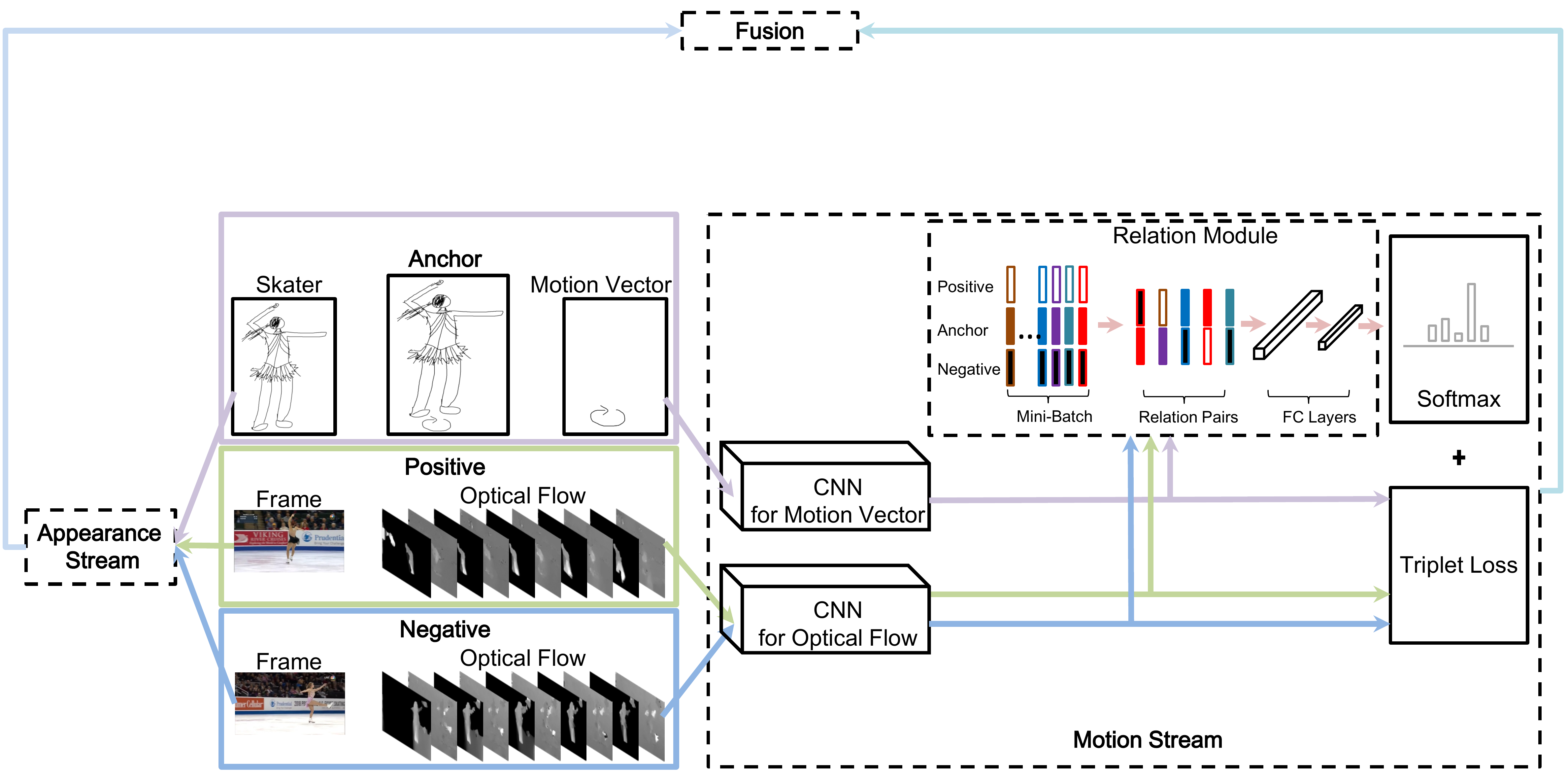}
	\caption{Architecture of the proposed framework for FG-SBVR. Best viewed in color.}
	\label{fig:pipeline}
\end{figure*}

\noindent\textbf{Motion Stroke Annotation}\quad~While drawing each sketch, the volunteers annotate  which strokes form the motion vector. This simplifies separation of  the ``skater'' and ``motion'' components of each sketch for our proposed multi-stream multi-modality model~(see the sketch input part in Figure~\ref{fig:pipeline}).

\noindent\textbf{Sketch-to-Frames Annotation}\quad To provide strong supervision for correspondence between multi-page sketches and videos, the volunteers next annotate which video frames correspond to each of their sketches. This correspondence annotation is illustrated in Figure~\ref{fig:film}. We will explore the importance of using this information later (see Section~\ref{sec:experiments}).


\cut{
\noindent\textbf{Motion Type Annotation}\quad Finally,  annotations with respect to the motion of sketches are collected. Although the motion vectors are drawn at the fine-grained instance level, we integrate and group all the sketches into $37$ classes~(including static sketch without motion vector) according to their approximate directions and angles. Then, a unique label is assigned to each motion class. Therefore, each video clip has a motion label sequence to describe its motion pattern, which is shared with its corresponding sketch sequence. This can also be used as supervision when training our model.
\textcolor{red}{This paragraph should be deleted!}
}

\begin{figure*}[!htp]
	\centering
		\includegraphics[width=0.75\textwidth]{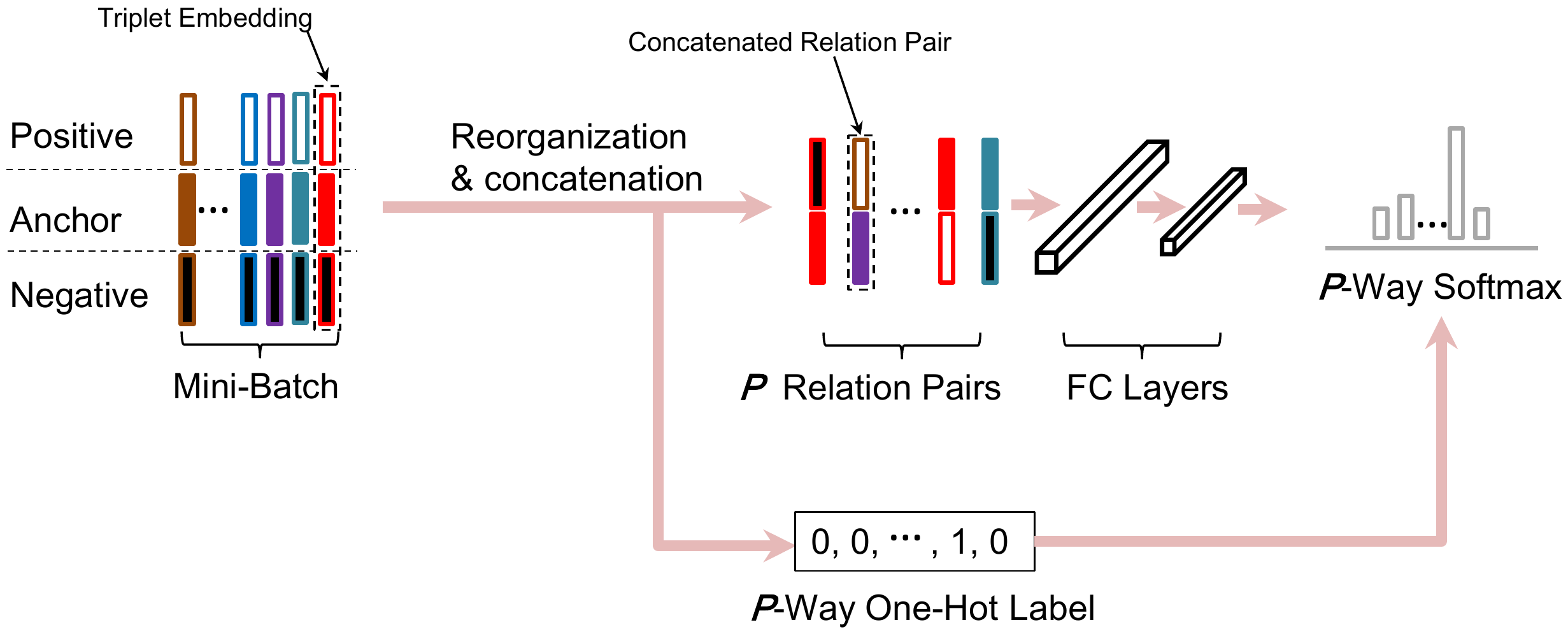}
	\caption{Detailed illustration of our relation module. Best viewed in color.}
	\label{fig:relation_module_architecture}
\end{figure*}

\section{Methodology}

\noindent\textbf{Problem Setting} We assume that the training dataset $D$ consists of $N$ paired sketch sequences and video clips: $D=\{(S_i,V_i)\}^N_{i=1}$. Each sketch sequence $S_i$ is composed of $M_i$ sketch `pages', and each of these has an appearance and motion component: \textcolor{black}{$S_i=\{(s^{ap}_j,s^{mo}_j)\}^{M_i}_{j=1}$}. Similarly, each video clip $V_i$ is composed of $O_i$ frame chunks: \textcolor{black}{$V_i=\{(v^{ap}_j,v^{mo}_j)\}^{O_i}_{j=1}$}. Given $D$, we aim to learn a deep sketch and video mult-modal joint embedding space, where the similarity of a sketch query and video pair can be simply computed as a distance for retrieval.


\subsection{Model}\label{sec:two-branch-cross-modal}

{\fgsbvr} must account for both appearance and motion in  fine-grained matching. Therefore,  we develop a multi-stream multi-modality joint embedding framework for {\fgsbvr} that processes appearance and motion using two streams respectively for both modalities. Subsequently, we combine the complementary representations extracted from these two streams to obtain a fused representation. Similarly to several fine-grained retrieval applications \cite{yu2016sketch,sangkloy2016sketchy}, we use triplet loss \cite{schroff2015facenet} to supervise training. In addition, a softmax cross-entropy loss is added to the \moduleName{} sub-network that labels multiple matching and mismatching pairs within one mini-batch.
Our model architecture is illustrated in Figure~\ref{fig:pipeline}, where the four components: Input, Appearance Stream, Motion Stream, and Fusion Mechanism, as well as their relationships are shown. We next detail each of these components.

\vspace{0.1cm}\noindent\textbf{Model Input}\quad Within the triplet loss paradigm, training tuples are constructed and each includes: sketch anchor, positive (matching) video, negative (mismatching) video. 
\cut{Each sketch \doublecheck{$s$ is separated into ``skater''~sketch $s^{skater}$~and ``motion vector'' sketch~$s^{motion}$}, which are inputted into the appearance and motion stream respectively as anchors. For sketches depicting static posture without motion, we use a blank sketch as the motion input.}
We explore various triplet construction strategies which will be detailed later. For a given triplet, our operations are: Firstly, we choose one video frame~\textcolor{black}{from the positive video} as the positive atom of the triplet for appearance stream. Secondly, use this selected frame as a start frame to compute $L$ pairs of consecutive optical flows\cut{ ($L$ is an empirical hyper-parameter)} using the GPU implementation of TV-$L^1$ \cite{zach2007duality} in OpenCV.
Following~\cite{simonyan2014two}, we calculate the optical flow matrix pair along $x$ and $y$ directions and alternately stack them to form a total of $2L$ input channels.
This $2L$ channels of optical flow will work as the positive atom of triplet for motion stream. Thirdly, the same operations are applied to negative atoms for the two streams. 
\textcolor{black}{Finally, the sketch anchor is separated into ``skater'' appearance~$s^{ap}$ and ``motion vector''~$s^{mo}$ components, which are then used in  the appearance stream and motion stream as anchors, respectively.}
Therefore, as shown in Figure~\ref{fig:pipeline}, there are six input branches in total.

\vspace{0.1cm}\noindent\textbf{Appearance Stream CNN}\quad
During training, there are three appearance branches corresponding to the atoms of a triplet $t\textcolor{black}{^{ap}}=(s^{ap}, v^{ap,+}, v^{ap,-})$. We assume weight sharing between all three branches (positive and negative frame, appearance sketch).
The appearance stream backbone is  GoogLeNet Inception V$3$~\cite{szegedy2016rethinking}. The loss for triplet $t^{\textcolor{black}{ap}}$ is:
\begin{equation}
\label{equ:triplet_loss}
\begin{split}
\mathcal{L}_{t} = \max (0, \triangle &+ \|{\mathcal{F}}_{\Theta_{ap,c}}(s^{ap}) - {\mathcal{F}}_{\Theta_{ap,c}}(v^{ap,+})\|_2^2 \\ 
&- \|{\mathcal{F}}_{\Theta_{ap,c}}(s^{ap}) - {\mathcal{F}}_{\Theta_{ap,c}}(v^{ap,-})\|_2^2),
\end{split}
\end{equation}
where~$\triangle$ is a margin and ${\mathcal{F}}_{\Theta_{ap,c}}$ denotes CNN feature extraction by the appearance stream network, parameterised by $\Theta_{ap,c}$.

\vspace{0.1cm}\noindent\textbf{Motion Stream}\quad
Similar to the appearance stream, there are three branches corresponding to triplet atoms~$t^{mo}=(s^{mo}, v^{mo,+}, v^{mo,-})$ during training. Due to the different numbers of input channels in sketch and stacked optical flow fields~($3$ vs. $2L$), we use two CNNs with different input depth. Both are GoogLeNet Inception V$3$-based, but the latter is modified to use $2L$ input channels. The loss is also a triplet loss combined with a relation loss analogous to Equation~\ref{equ:total_loss}.
We use ${\mathcal{F}}_{\Theta_{mo,c}}$ to denote CNN feature extraction by the motion stream network, parameterised by $\Theta_{mo,c}$.

\vspace{0.1cm}\noindent\textbf{\ModuleName{}}\quad Limited
training sketch-video pairs make a \fgsbvr~model vulnerable to overfitting and poor generalisation to test data. Here we introduce our \moduleName{}  ~\cite{sung2018learning,xie2018comparator} 
\textcolor{black}{that can be independently applied to both our appearance stream and motion stream
to alleviate the data scarcity problem.} The idea is that, instead of modelling triplet ranking relationships, we form larger groups and model the more complex group relationship. This aims to maximise the use of the limited training data, because when the group size is larger than 3, there can be many more groups than triplets. 
\textcolor{black}{
More specifically, given a mini-batch,
 we forward it through our appearance stream or motion stream CNN, and obtain a mini-batch of triplet embedding vectors. 
Then, we randomly select and reorganise $P$ sketch-video relation pairs, forming one true match pair and $(P-1)$ false match pairs. As shown in Figure~\ref{fig:relation_module_architecture}, we concatenate the embedding vectors for each relation pair, and input it into a relation  network consisting of two fully connected layers. 
We will set the dimensionality of our CNN output embedding as $256D$ in this paper, so that the input dimensionality of our relation module will be $512D$.
Simultaneously, the associated ground-truth pairwise relationships are formed as a $P$-dimension one-hot vector as training objective, in which the non-zero element corresponds to the true match pair. 
We adopt $P$-way cross-entropy softmax loss for our \moduleName{} sub-network, denoted as ``relation loss'' $\mathcal{L}_{r}$. $P$ is set to 5 in this work. Thus, the total loss for each mini-batch can be defined as}
\begin{equation}
\label{equ:total_loss}
\begin{split}
\mathcal{L} = \mathcal{L}_{t} + \lambda_1 \mathcal{L}_{r},
\end{split}
\end{equation}
where  $\lambda_1$ is a weighting factor.

\textcolor{black}{
In particular, in this work, we independently train a relation module for each stream, parameterised by $\Theta_{ap,r}$ and $\Theta_{mo,r}$ respectively. This is to say that the loss function of each stream contains both a triplet term and a relation loss.
The detailed training and optimization are described in Algorithm~\ref{alg:1}.
}

\vspace{0.1cm}\noindent\textbf{Stream Fusion}\quad Once the triplet ranking and \moduleName{} training for the two streams is complete, a natural way to combine them is to concatenate or fuse them with another FC layer, and fine-tune with another triplet loss. However, similarly to the observation in~\cite{simonyan2014two}, this fails in our case due to overfitting. 
Thus, we use two fusion approaches to fuse our two streams: (i) ranking-based fusion, and (ii) feature concatenation fusion. For ranking-based fusion, a sketch query $\textcolor{black}{S}$ generates a ranked list of matching videos $\{\textcolor{black}{V}_j\}$ based on \textcolor{black}{E}uclidean distance. We use $r^{ap}_j$ and $r^{mo}_j$ to indicate the rankings of each video $j$ using the appearance and motion stream features respectively. The final ranking $r_j$ of each gallery video clip is the weighted arithmetic mean of its appearance and motion ranks:
\begin{equation}
\label{equ:weighted_arithmetic_mean}
\begin{split}
r_j = \lambda_2 r_j^{ap} + (1 - \lambda_2) r_j^{mo},
\end{split}
\end{equation}
where $\lambda_2$ is the weighting factor.
For the feature concatenation strategy, we concatenate the features from two streams as the final representation, and then conduct Euclidean distance based ranking.

\vspace{0.1cm}\noindent\textbf{Training with Strong Supervision}\quad Recall that our sketch queries can contain multiple pages corresponding to different segments/sub-clips within the video clip, and that the detailed correspondence is annotated. This provides the strongest supervision for our task. Specifically, for each single sketch  anchor, its positive video candidates are frames within the corresponding sub-clip. Frames outside the corresponding sub-clip, or frames in different clips entirely, are treated as negative. 

\begin{algorithm}[!t]
    	\caption{Learning algorithm for our proposed multi-stream multi-modality FG-SBVR deep network.}
  	\label{alg:1}
        \begin{algorithmic}
        \Require  $D=\{(S_i,V_i)\}^N_{i=1}$.
        \State 1. Train appereance stream as following loop.
        \For{number of training iterations}
            \State 1.1 Forward mini-batch through CNN, calculate $\mathcal{L}_{t}$.
            \State 1.2 Reorganise mini-batch and forward relation pairs through relation module, calculate $\mathcal{L}_{r}$.
            \State 1.3 Update $\Theta_{ap,c}$ using $\mathcal{L}_{t}$ and $\mathcal{L}_{r}$.
            \State 1.4 Update $\Theta_{ap,r}$ using $\mathcal{L}_{r}$.
        \EndFor
        \State 2. Train motion stream as following loop.
        \For{number of training iterations}
            \State 2.1 Forward mini-batch through CNN, calculate $\mathcal{L}_{t}$.
            \State 2.2 Reorganise mini-batch and forward relation pairs through relation module, calculate $\mathcal{L}_{r}$.
            \State 2.3 Update $\Theta_{mo,c}$ using $\mathcal{L}_{t}$ and $\mathcal{L}_{r}$.
            \State 2.4 Update $\Theta_{mo,r}$ using $\mathcal{L}_{r}$.
        \EndFor
        \Ensure CNN embedding extractions ${\mathcal{F}}_{\Theta_{ap,c}}$ and ${\mathcal{F}}_{\Theta_{mo,c}}$.
        \end{algorithmic}
    \end{algorithm}

\subsection{Weakly Supervised Learning}
\label{sec:WSL}
The conventional strongly supervised setting requires labor intensive cross-modal annotation (Figure~\ref{fig:film}). It would be easier to scale {\fgsbvr} if the model could be learned with less detailed and intensive annotation.  Weakly supervised learning (WSL) for {\fgsbvr} is thus required. This can be formulated as a multi-instance learning (MIL) \cite{andrews2003support} problem. Given an  anchor sketch-page, there will be video clips to which it definitely does not correspond (negative bag), and those to which it will correspond to at least one frame within the clip (positive bag). Each of those clips (bags) contains multiple frames (instances). In particular, we consider that we have the sketch-sequence to video-clip pairing, but not the detailed page-frame level pairing (Figure~\ref{fig:film}, green lines). In this case, there is one positive bag per anchor page, \ie, the corresponding clip. All non-corresponding clips are negative bags. The challenge is to correctly estimate the positive/matching, and negative/mismatching instances (frames) within each positive bag (video).

\noindent\textbf{Pruning Heuristics and Initialisation} 
\quad \cut{We use some heuristics to prune the multi-instance learning search space. During our data collection (Section~\ref{sec:data_annotation}) sketches are annotated with skater identity, page-number within the sketch sequence, and motion types are clustered.} 
To prune the search space of positive bags defined above, we further consider: (i) If a query sketch is the first in a sequence, its  positive bag consists of the first 50\% of frames in a potentially matching video according to the criteria above. (ii) If a query sketch is last in a sequence, its positive bag consists of the last 50\% frames in a potentially matching video. (iii) Otherwise the positive bag is the middle 50\% of clip frames. Both appearance and motion streams share the same criteria. All instances (frames) within positive bags are initialised as positive. 

\noindent\textbf{Multi-Instance Learning}\quad
Given the above bag definition and initialisation, we iteratively refine the labels of instances within positive bags using multi-instance training. As per conventional MIL  \cite{andrews2003support}, training alternates between phases of classifier/representation learning, and re-estimation of positive instances as follows:
(i) Learning: Update the network with back-propagation and loss $\mathcal{L}$, assuming the positive/negative instance labels are fixed.
(ii) Label Inference: Set the least likely matches (furthest $T\%$  distance) frames in positive bags to be negative, assuming the network parameters are fixed.



\begin{figure}[!t]
	\centering
		\includegraphics[width=\columnwidth]{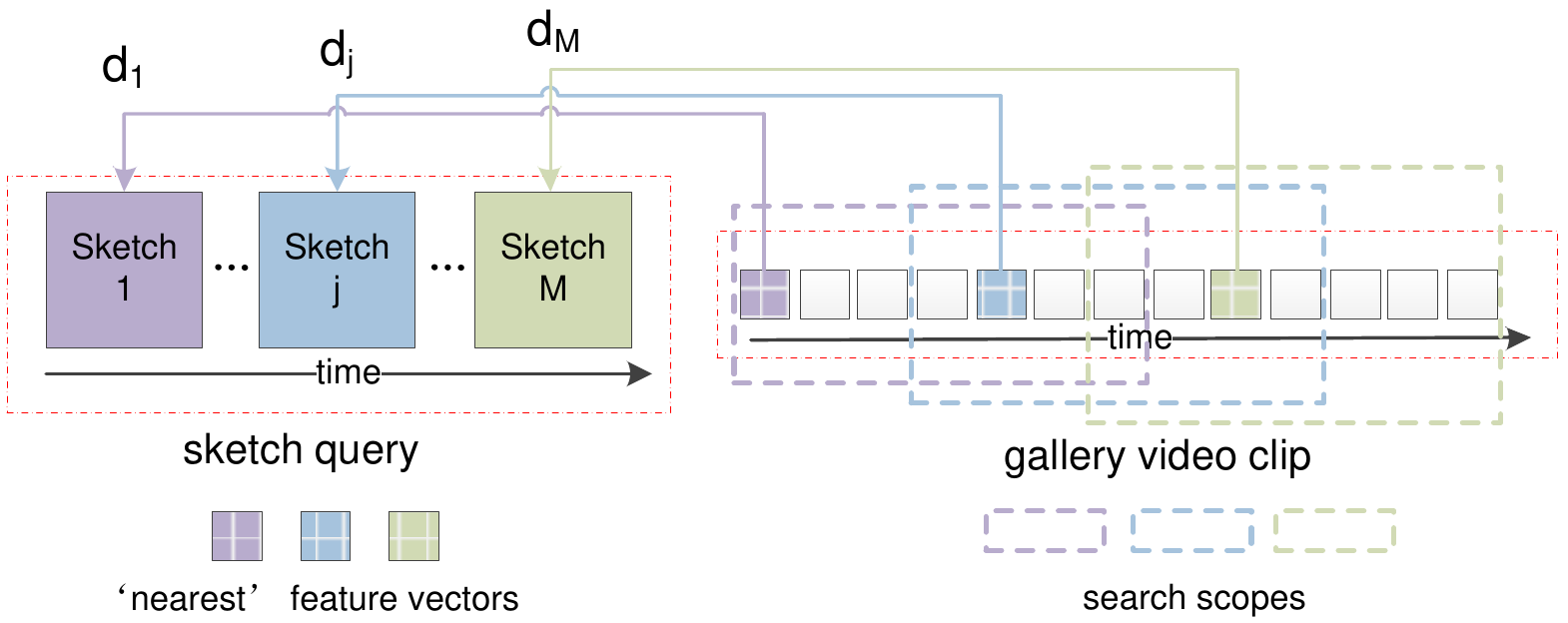}

	\caption{Illustration for the similarity distance calculation for sketch query and gallery video. Best viewed in color.}
	\label{fig:temporal_test}
\end{figure}

\subsection{Model Deployment}

Given a trained model, during testing we need to match a sketch-sequence to a video clip (frame-sequence). However, our network has learned to rank sketch-pages and video-frames. Thus, we need to define how to aggregate a set of pairwise page-frame scores. 
\textcolor{black}{In testing, we use the CNN output embedding to conduct retrieval.}
In particular, after deep feature extraction by our two-stream network, sketches and videos are respectively represented as: $S=\{\textcolor{black}{s}_j\}^M_{j=1}=\{(F_{\Theta_{ap,c}}(s_j^{ap}),F_{\Theta_{mo,c}}(s_j^{mo})\}^M_{j=1}$ and $V=\{\textcolor{black}{v}_k\}^O_{k=1}=\{(F_{\Theta_{ap,c}}(v_k^{ap}),F_{\Theta_{mo,c}}(v_k^{mo})\}^O_{k=1}$ respectively. A simple solution is to choose the match that has the lowest sum of nearest neighbour matching costs:
\begin{equation}
\label{equ:distance}
\begin{split}
D(\textcolor{black}{S},\textcolor{black}{V}) = \frac{1}{M}\sum^M_{j=1}{\min_{k \textcolor{black}{\in [\varphi(j), \psi(j)]}} d(\textcolor{black}{s}_j,\textcolor{black}{v}_k)},\\
\tiny
\textcolor{black}{
(\begin{cases}
\varphi(j)=1~~and~~\psi(j) = O/2, & j=1\\
\varphi(j)=O/2~~and~~\psi(j) = O, & j=M\\
\varphi(j)=O/4~~and~~\psi(j) = 3O/4,& other
\end{cases}),}
\end{split}
\end{equation}
\textcolor{black}{where $\varphi(j)$ and $\psi(j)$ are lower and upper bounds of $k$.}
Figure~\ref{fig:temporal_test} provides an illustration.

\cut{However, this strategy does not exploit any temporal sequence information. Therefore we propose the following simple heuristic to perform matching in a temporal-sequence-aware way. 
\todo{Peng: Insert a paragraph to explain it properly here!} }

\begin{table*}[h]
\caption{Ablative evaluation of our {\fgsbvr} model under strong supervision.  
Chance level performance is $0.0078$~($\approx 1 / 128$) acc.@1.}
\label{table:retrieval_results}
\small
\begin{center}
\begin{tabular}{ c  c c c c c c c}
\hline
  &\multicolumn{3}{c}{Triplet Ranking} & &  \multicolumn{3}{c}{Triplet Ranking + \ModuleName{}}  \\
\cline{2-4}  \cline{6-8}
    Model     & acc.@1 & acc.@5 & acc.@10 &  & acc.@1 & acc.@5 & acc.@10 \\

\hline
 app. stream  & 0.1250 & 0.3281 & 0.4063 & & 0.1719  & 0.3516 &  0.5156\\
                                     motion stream      & 0.1406 & 0.3438 & 0.5469 & &  0.1719  & 0.3828  &  0.5781\\
                                    ranking fusion       & 0.2188 & 0.4141 & 0.5547 & &  \textcolor{black}{0.2969}  &  0.6094  &  0.7344 \\
                                    concat fusion  &       0.3047 & 0.6172 & 0.7344 & &  \textcolor{black}{0.3438} &  0.6094  &  0.7656 \\
                            \hline
\end{tabular}
\end{center}

\end{table*}

\begin{table*}[th]
\caption{{\fgsbvr} retrieval results obtained with weak supervision. }
\label{table:retrieval_results_weak_supervision}
\small
\begin{center}
\begin{tabular}{ c  c c c c c c c}
\hline
  &\multicolumn{3}{c}{Triplet Ranking} & &  \multicolumn{3}{c}{Triplet Ranking + \ModuleName{}}  \\
\cline{2-4}  \cline{6-8}
    Model     & acc.@1 & acc.@5 & acc.@10 &  & acc.@1 & acc.@5 & acc.@10 \\

\hline
 app. stream  & 0.0234 &	0.1016 &	0.1719 & & 0.0469	& 0.1094 &	0.1641 \\
                                     motion stream      & 0.0469 &	0.1250 &	0.2500 & & 0.0703 &	0.1797 &	0.2656\\
                                    ranking fusion       & 0.0547 &	0.1094 &	0.1641 & & 0.0859 & 0.1172 & 0.1719 \\
                                    concat fusion        & 0.0625 &	0.1094 &	0.1953 & & 0.0703 & 0.1250 & 0.1875 \\
                            \hline
\end{tabular}
\end{center}

\end{table*}



\begin{table}[t]
\caption{{\fgsbvr} results of 3D CNN based baselines under strong supervision setting.}
\label{table:baselines_retrieval_results}
\small
\begin{center}
\resizebox{\columnwidth}{!}{
\begin{tabular}{c c c c c}
\hline

3D CNN  &    Model   & acc.@1 & acc.@5 & acc.@10 \\
\hline
\multirow{4}{*}{Non-Local~\cite{wang2018non}} & app. stream  &  0.0938 &	0.2109	& 0.3359 \\
                                    & motion stream      & 0.0703	& 0.1953	& 0.3047  \\
                                    & ranking fusion       &   0.1016 &	0.1875	 & 0.3281  \\
                                    & concat fusion        &   0.1016 &	0.2969 &	0.4141  \\
\hline
\multirow{4}{*}{ \tabincell{c}{3D ResNet18~\cite{hara2018can}}} & app. stream &  0.0469 &	0.0781 & 0.1094 \\
                                    & motion stream     &  0.0469 &	0.1016 &	0.1563 \\
                                    & ranking fusion      &  0.0234 & 0.0781 &	0.1250 \\
                                    & concat fusion       &  0.0547 & 0.0859 &	0.1328 \\
\hline

\multirow{4}{*}{T-C3D~\cite{liu2018t}} & app. stream &  0.0391 & 0.1094 & 0.1719 \\
                                    & motion stream     &  0.0234 &	0.0938 & 0.1641 \\
                                    & ranking fusion      &  0.0313 &	0.0859 & 0.1484 \\
                                    & concat fusion       &  0.0313 &	0.0703 &	0.1719 \\

\hline

\end{tabular}
}
\end{center}

\end{table}

\begin{table}[t]
\caption{Performance improvement on Non-Local Neural Network~\cite{wang2018non} achieved by introducing the \moduleName.  }
\label{table:non-local meta}
\small
\begin{center}
\begin{tabular}{ c c c c}
\hline

  Model   & acc.@1 & acc.@5 & acc.@10 \\
\hline
app. stream  &  0.0235 &	0.0781	& 0.0703 \\
motion stream      & 0.0156	& 0.0234	& 0.0313  \\
ranking fusion       &   0.0078 &	0.0235	 & 0.0000  \\
concat fusion        &   0.0157 &	0.0625 &	0.0313  \\
\hline
\end{tabular}
\end{center}

\end{table}

\section{Experiments}
\label{sec:experiments}

\subsection{Experiment Settings}

\noindent\textbf{Dataset Split}\quad We generate training, validation, and testing sets by randomly splitting the 528 clips into 350 (with $971$ sketches) for training, 50 (with 131 sketches) for validation, and 128 (with 346 sketches) for testing. Thus during testing, we have 128 sketch sequences and 128 video clips as the queries and gallery, respectively.

\vspace{0.1cm}\noindent\textbf{Implementation Details}\quad All experiments are implemented in PyTorch, and run on a single TITAN Xp GPU. We use model hyperpameters values obtained using 5-fold cross-validation on the training set: $\delta=0.5,L=5,T\%=0.1, \lambda_1 = 0.001,\lambda_2 = 0.5 $.  RMSprop optimizer is used with initial learning rate  $0.001$ and mini-batch size is $16$.
We initialise our branches with Inception V3 with the following modifications: The original ``fc'' components of Inception are replaced by two new fully-connected layers ($2048D \rightarrow 512D$, $512D \rightarrow 256D$). For the motion stream, the first layer with non-standard depth is randomly initialised. 
Our \moduleName{} sub-network has two fully-connected layers~($512D \rightarrow 128D$, $128D \rightarrow 32D$).
The input size of the appearance \textcolor{black}{stream} and sketch motion \textcolor{black}{sub-}stream is $3 \times 299 \times 299$. The optical flow sub-stream input is \cut{$3 \times 299 \times 299$ and }$10 \times 299 \times 299$.

\vspace{0.1cm}\noindent\textbf{Evaluation Metric}\quad Given our instance-level retrieval task, we use retrieval accuracy as a metric. We quantify this by cumulative matching accuracy at various ranks. Acc@K is the proportion of sketch sequence queries whose true-match video clips are ranked in the top K (1 means 100\%).

\vspace{0.1cm}\noindent\textbf{Competitors}\quad
FG-SBVR is a brand new problem, thus no existing methods can be compared directly here. The main competitors here are the state-of-the-art video analysis models extended for the FG-SBVR problem. More specifically, for fair comparison, we still employ a multi-stream multi-modality network. But both streams of the video modality are now based on the latest 3D convolutional networks including Non-Local Neural Network~\cite{wang2018non}, 3D ResNet18~\cite{hara2018can}, 
and  T-C3D~\cite{liu2018t}. The two sketch streams are unchanged and aligned with the duplicated video streams in the hope that the static and dynamic aspects of the video contents can be disentangled through the alignment process, trained with the same triplet and \moduleName{} losses.  In addition, for an ablation study,  we also compare several variants of our own model. In particular, we compare our model trained with different types of supervision, including strong supervision (see Section~\ref{sec:two-branch-cross-modal}) and weak supervision (see Section~\ref{sec:WSL}). Since our model has static appearance and dynamic motion streams, we also compare variants of our model with one stream only and the full model with the two-streams fused in different ways (see Section~\ref{sec:two-branch-cross-modal}). Note that we also attempted to put RNN on top of the video and sketch streams to explicitly encode the temporal order information following \cite{BiDiRNN2016}, but the results are much worse so not reported here. 

\subsection{Results}
\label{sec:results}

\noindent\textbf{Strong Supervision}\quad 
Under strong supervision,  we first conduct an ablative study on the contributions of the multi-stream multi-modal architecture design and \moduleName{} in our model. The following observations can be made from Table~\ref{table:retrieval_results}: 
(i) In terms of single-stream performance, motion  outperforms appearance particularly at higher ranks. This is interesting: Each motion vector sketch component contains only 1 or 2 strokes, whilst the mean number of strokes for the skater sketch component is around 100 (see Table \ref{table:datasets_comparison}). Those 1-2 motion strokes seem worth 100 strokes that are used for depicting static appearance of the skater and her representative posture. 
(ii) When the two streams are fused using either ranking   or feature concatenation,  the performance is improved significantly. These results confirm that  appearance and motion patterns contain complementary information for {\fgsbvr}. Further, among the two fusion strategies, feature concatenation is clearly more effective. 
\textcolor{black}{(iii) When the \moduleName{} is added,  large improvements on retrieval accuracy are obtained. }  

Next, we compare our full model with a number of baselines extended from  state-of-the-art 3D CNN models. Table \ref{table:baselines_retrieval_results} shows the results obtained by three models based on 3D CNNs introduced in 2018 deployed in the same multi-stream multi-modality network, trained with the same triplet ranking and \moduleName{}. Comparing Table \ref{table:baselines_retrieval_results} with Table~\ref{table:retrieval_results}, it is clear that our model with the video modality decomposed explicitly into dynamic and static parts, modelled with optical flow and image CNN respectively, is much better (24\% higher on acc.@1). These results thus further validate our model design.  
\textcolor{black}{It is also interesting to note that for non-local network~\cite{wang2018non}, stream fusion helps and concatenation based fusion is the most effective strategy.}

\begin{table*}[!t]
\caption{Impact of number of pairs per mini-batch used by \moduleName.}
\label{table:comparison_on_meta_learning_pairs}
\small
\begin{center}
\begin{tabular}{c c c c c}
\hline

    model    & \tabincell{c}{Num paris \\ per \\ mini-batch}  & acc.@1 & acc.@5 & acc.@10 \\
\hline
strong supervision app. stream~(SAS) & 5 & 0.1719  & 0.3516 &  0.5156 \\
strong supervision app. stream~(SAS) & 10 &  0.1953	& 0.3672	& 0.5234  \\
strong supervision app. stream~(SAS) & 15 & 0.2031 &	0.4531 &	0.5859 \\
strong supervision app. stream~(SAS) & 20 &  0.2344 &	0.5156 &	0.6094 \\


\hline
\end{tabular}
\end{center}

\end{table*}

\keypoint{\ModuleName{} Analysis} To understand the impact of our \moduleName, we also study its efficacy in combination with the baselines. Specifically, we pick the strongest baseline, Non-Local network~\cite{wang2018non} and compute the performance difference between the models with and without the \moduleName. Table \ref{table:non-local meta} shows that in most cases, the \moduleName brings improvement, indicating the general applicability of such meta-learning inspired techniques for dealing with scarce training data. 

One reason for the efficacy of our \moduleName{} is that it leverages more negative pairs within each mini-batch. Table \ref{table:comparison_on_meta_learning_pairs} analyses the \moduleName{} from this perspective, showing that increasing the relation pairs per mini-batch leads to improved retrieval  performance. For our main results in this paper, we use $5$ relation pairs within each mini-batch due to limited GPU memory.  




\begin{table}[t]
\caption{Sketch-based action detection results. Chance performance (acc.@1) is $0.0029$~($\approx 1 / 346$).}
\label{table:action_detection_results}
\small
\begin{center}
\begin{tabular}{ccp{0.8cm}<{\centering}}
\hline
Supervision  &    Model      & Accuracy \\
\hline
\multirow{4}{*}{\tabincell{c}{Strong \\Supervision\\}} & app. stream  & 0.4133 \\
                                    & motion stream     & 0.3382  \\
                                    & ranking fusion       & 0.4682 \\
                                    & concat fusion       & \textcolor{black}{0.4798} \\


\hline


\multirow{4}{*}{\tabincell{c}{Weak \\Supervision\\}} & app. stream  & 0.1185  \\
                                    & motion stream    & 0.1936 \\
                                    & ranking fusion       & 0.1850 \\
                                    & concat fusion       & 0.1647 \\

\hline
\end{tabular}
\end{center}

\end{table}

\vspace{0.1cm}\noindent\textbf{Weak Supervision} \quad
Table~\ref{table:retrieval_results_weak_supervision} also shows the results obtained when our model is trained with weak supervision.  It is clear that: (i) Like the strong supervision setting, single motion stream outperforms single appearance stream.  (ii) Two-stream fusion now does not guarantee  to improve performance, due to the poor performance of the appearance stream.  (iii) All the accuracy values are significantly lower than that under strong supervision, as expected.   Overall, it is worth pointing out that there is a large scope for further improvement under this challenging setting, in order to narrow the distinct performance gap between supervised and weakly supervised models.

\vspace{0.1cm}\noindent\textbf{Fine-Grained Sketch-Based Action Detection}\quad Going beyond FG-SBVR, we can also use our method and dataset to explore an even more fine-grained application, namely fine-grained instance-level sketch-based action detection~(FG-SBAD).  Given a motion sketch-page and a video clip, the goal of FG-SBAD is to localise the target action depicted by the sketch. 
We propose a straightforward solution to FG-SBAD: traverse video clip frame-by-frame and report the index of the nearest-neighbour frame. If the proposal is in the range (\ie, within 5 frames) of the \textcolor{black}{sketch-to-frames ground truth}, then it is regarded as a successful detection.  As illustrated in Table~\ref{table:action_detection_results}, the strongly-supervised concatenation fusion approach performs much better than the  weakly supervised alternative. 

\begin{figure*}[!thp]
	\centering
		\includegraphics[width=\textwidth]{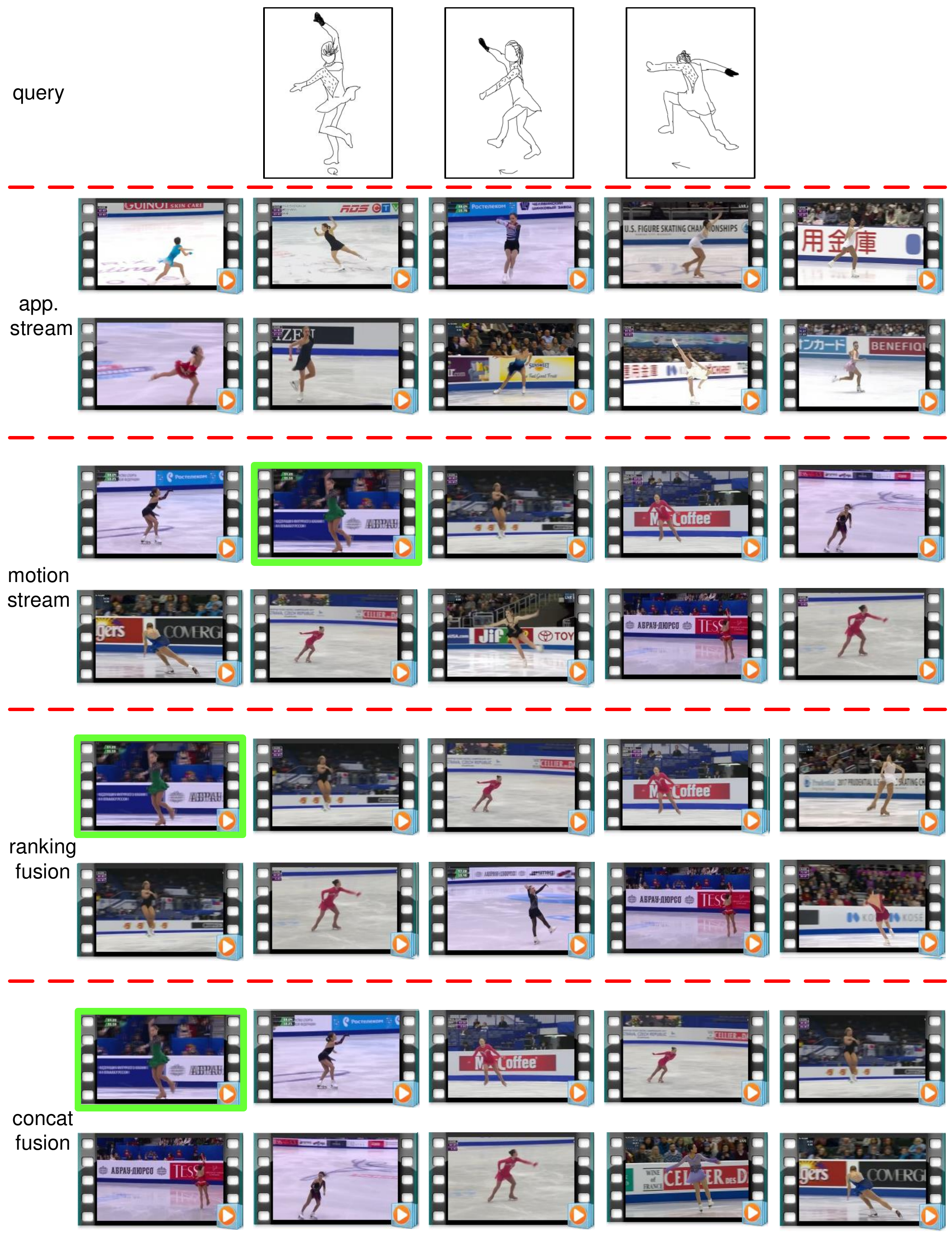}

	\caption{Qualitative comparison of top $10$ retrieval results using different variants of our models under the strongly supervised setting.  The 10 videos are ordered from top to bottom and from left to right according to their ranks. The true matches are highlighted in green.}
	\label{fig:retrieval_visualization}
\end{figure*}

\begin{figure*}[!thp]
	\centering
		\includegraphics[width=\textwidth]{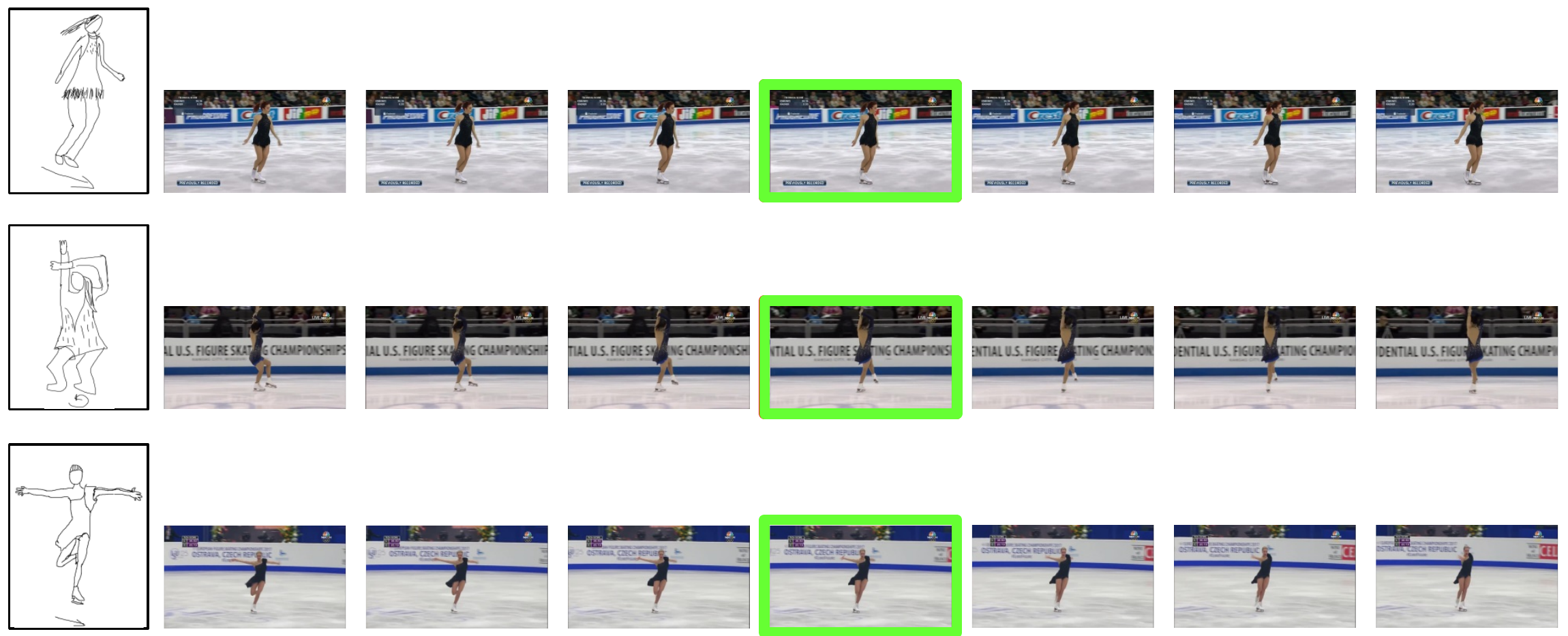}

	\caption{Results for sketch-based action detection of our concatenation fused full model under the strongly supervised setting. In each row, sketch is on the left, and the green-bordered frame is best matched frame. The neighbouring six frames are also shown.}
	\label{fig:visualization_4_action_detection}
\end{figure*}

\noindent\textbf{Qualitative Results}\quad
We next show some visual examples of the retrieval results obtained using  our multi-stream
multi-modality model and its variants.   In Figure~\ref{fig:retrieval_visualization}, a sequence of three sketches are used as query and the top-10 ranked videos using different models are shown. The query  sketch sequence captures the key moments of the video sequence. In particular, the motion vector parts of the three sketches indicate one spin movement and two glissade movements, respectively; in the meantime, the skater parts of the sketches contain visual details of the skater's appearance such as  stripes on the clothes and glove, as well as her body posture at those key moments.  From the retrieval results, it can be seen that: (i) The multi-stream fusion models (feature concatenation based fusion and ranking based fusion) give the desired results -- the true match is ranked at the top.  (ii) From the retrieval results of the appearance-stream model variant, we can see that although the correct match is not in the top 10,
all skaters in the top~$10$ retrieved videos wear gloves or single shoulder dress, similar to those of the skater in the true match video. These results suggest that without the motion vector, the skater part of the sketches is not discriminative enough for the model to find the correct skating sequence -- the model put too much emphasise on the static appearance of the skater rather than her movements.  (iii) By contrast, the motion-stream model variant is able to retrieval video sequences containing similar spin or glissade movements with the given query. However, without any information about the static appearance of the skater, the model is unable to distinguish video sequences of similar skating routines but performed by different skaters.  

\noindent\textbf{Localisation}\quad
Note that our two-stream model is able to produce a similarity/matching score between a query sketch and each frame of a video sequence (\ie, Fine-Grained Sketch-Based Action Detection as discussed in Section 5 of the main paper). In Figure~\ref{fig:visualization_4_action_detection}, we show that given a query sketch, which frame in the correctly matched video sequence has the highest matching score. The results suggest that, if the video sequence can be correctly retrieved, our model can be used to accurately localize which specific time of the sequence the query sketch is depicting. In particular, the body pose of the skater in the best matched frame is remarkably  similar to the body pose of the sketched skater.

\noindent\textbf{Running Cost}\quad All our experiments are conducted on an Intel Core i7-7700K $4.2$ GHz CPU and single TITAN Xp GPU. Training our two-stream model with  strong supervision and weak supervision takes about $20$ hours and $30$ hours, respectively. For testing, it takes about $200$ milliseconds for one retrieval.

\section{Conclusion and Future Work}

We introduced the novel task of fine-grained instance-level sketch-based video retrieval (FG-SBVR) and a dataset to enable the research on this task.  We also  proposed a novel multi-stream multi-modality network with \moduleName{} to solve this problem in both strongly- and weakly-supervised settings. Our new dataset can also support future research on tasks such as  video summarisation, sketch-based video generation, and  multi-modal tasks that combine visual and audio cues (via commentary track). 
For example, the dataset can be used directly for video summarisation to complement existing datasets such as TVSum \cite{song2015tvsum} and CoSum \cite{chu2015video}. The recent research on video-to-video synthesis \cite{wang2018vid2vid} can now be extended to cross-modal video-to-sketch and sketch-to-video synthesis using our dataset.

{\small
\bibliographystyle{IEEEtran}
\bibliography{egbib}
}

\end{document}